\documentclass[runningheads]{llncs}
\usepackage[T1]{fontenc}
\usepackage{graphicx}
\usepackage{amsmath}
\usepackage{wrapfig}
\usepackage{booktabs}
\usepackage{amsfonts}
\usepackage{svg}
\usepackage{amssymb}
\usepackage{xcolor}
\usepackage{colortbl}
\usepackage{subcaption}
\usepackage{multirow}
\usepackage{hyperref}
\usepackage{array}
\definecolor{firstplace}{RGB}{230, 241, 255}
\definecolor{secondplace}{RGB}{255, 241, 230}
\definecolor{thirdplace}{RGB}{240, 255, 240}

\begin{document}
\title{Explaining Vision GNNs: A Semantic and Visual Analysis of Graph-based Image Classification}
\titlerunning{A Semantic and Visual Analysis of Graph-based Image Classification}

\author{Nikolaos Chaidos\orcidID{0009-0006-0347-2785} \and
Angeliki Dimitriou\orcidID{0009-0001-5817-3794} \and
Nikolaos Spanos\orcidID{0009-0001-2691-0956}
\and
Athanasios Voulodimos \orcidID{0000-0002-0632-9769} \and Giorgos Stamou\orcidID{0000-0003-1210-9874}
}

\institute{National Technical University of Athens \\
\email{\{nchaidos, angelikidim, nspanos\}@ails.ece.ntua.gr} \\
\email{   thanosv@mail.ntua.gr
}
\email{gstam@cs.ntua.gr}\\}

\authorrunning{N. Chaidos et al.}

\maketitle     

\begin{abstract}
Graph Neural Networks (GNNs) have emerged as an efficient alternative to convolutional approaches for vision tasks such as image classification, leveraging patch-based representations instead of raw pixels. These methods construct graphs where image patches serve as nodes, and edges are established based on patch similarity or classification relevance. Despite their efficiency, the explainability of GNN-based vision models remains underexplored, even though graphs are naturally interpretable.  In this work, we analyze the semantic consistency of the graphs formed at different layers of GNN-based image classifiers, focusing on how well they preserve object structures and meaningful relationships. A comprehensive analysis is presented by quantifying the extent to which inter-layer graph connections reflect semantic similarity and spatial coherence. 
Explanations from standard and adversarial settings are also compared to assess whether they reflect the classifiers' robustness. 
Additionally, we visualize the flow of information across layers through heatmap-based visualization techniques, thereby highlighting the models' explainability. 
Our findings demonstrate that the decision-making processes of these models can be effectively explained, while also revealing that their reasoning does not necessarily align with human perception, especially in deeper layers. The code is available at \href{https://github.com/nickhaidos/Vision-GNNs-Explainer}{https://github.com/nickhaidos/Vision-GNNs-Explainer}.

\end{abstract}

\section{Introduction}

Image classification has long been dominated by powerful yet opaque models, raising questions about their decision-making processes. As these models achieve unprecedented accuracy, the need for transparency has become %ever% 
critical. To bridge this gap, techniques such as Grad-CAM \cite{Selvaraju2017GradCAM}, saliency maps \cite{simonyan2013deep}, and occlusion experiments \cite{zeiler2014visualizing} have provided insights into how image models prioritize image regions, illuminating the internal logic behind classification decisions. These methods have both advanced our grasp of deep models' inner workings and highlighted challenges in aligning machine explanations with human intuition.

\begin{wrapfigure}{r}{0.61\columnwidth}
% \vskip -0.2in
  \centering
  \includegraphics[width=0.63\columnwidth]{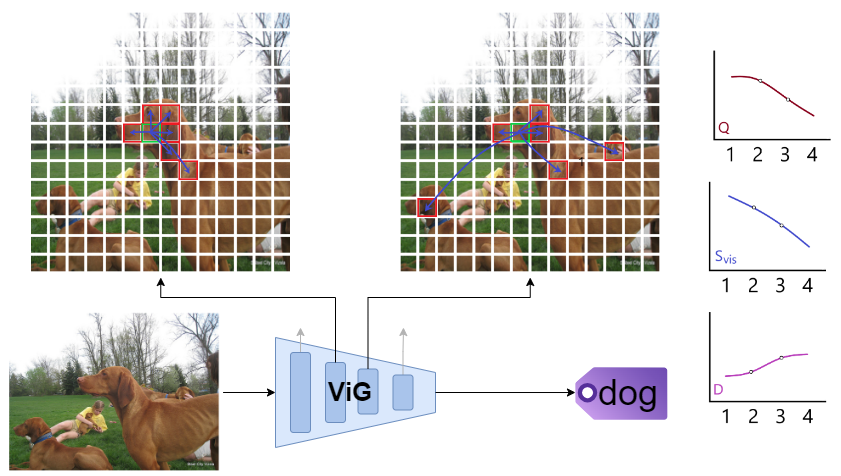}
  \caption{Visual depiction of ViG's representation of a "dog" image at different layers (2, 3). Subgraphs refer to the central green patch and its neighbors (red patches) at that layer. Diagrams represent object-based modularity ($Q$), visual similarity ($S_{vis}$) and spatial distance ($D$) with dots corresponding to the 2nd and 3rd layer. }
  \label{fig:teaser}
  \vskip -0.1in
\end{wrapfigure} 

Building on this rich history of interpretability, a new frontier is emerging with Graph Neural Networks (GNNs). Unlike Convolutional Neural Networks (CNNs) or Vision Transformers (ViTs) that operate on pixel grids, GNN-based approaches re-imagine images as collections of interrelated patches - nodes connected by edges that encode semantic or spatial similarity. This graph-based perspective has proven to be efficient thanks to %the% 
lightweight GNN models that ultimately process the graph representations of images for classification. Recent research has already explored several ways to further accelerate the process by focusing on the optimal construction of the graph - for example, by comparing dynamic and static graphs or applying directional constraints \cite{Munir2023MobileViGGS,Munir2024GreedyViGDA}.
Furthermore, claims have been made that GNN-based image classification can potentially provide sufficient interpretability for model decisions \cite{Han2022VisionGA} by offering explainability insights that mirror the way humans naturally perceive and organize visual information. However, their inner workings remain largely unexplored, leaving open questions about semantic consistency and the reliability of explanations across different layers.

The seminal approach by Han et al. \cite{Han2022VisionGA} (Vision GNN (ViG)) introduces the foundations of graph-based approaches, redefining image classification by structuring images as graphs rather than grids. In ViG, an image is first divided into non-overlapping patches that serve as nodes, while edges are formed based on spatial proximity or learned feature similarities.
Whereas CNNs mainly rely on local receptive fields, ViG propagates information through message passing, allowing for flexible and dynamic interactions between distant regions in the image. This approach not only reduces computational complexity but also captures richer contextual dependencies. 
Given its merits and its inherent inclination toward explainability, as hinted by its authors, ViG is a natural choice for a study on interpretability among other GNN-based vision models \cite{Han2023VisionHA,Munir2023MobileViGGS,Munir2024GreedyViGDA}.

To this end, in this paper we present the first explainability analysis of the ViG model, investigating whether its graph-based structure inherently leads to 
interpretable decisions. While graphs are often considered inherently interpretable - by providing a clear mapping of nodes to object parts and their relationships - we aim to rigorously evaluate this assumption both visually and numerically. Specifically, we examine whether patches belonging to the same object tend to connect with semantically similar or spatially close ones, illuminating the model’s inner workings.
Our approach leans toward a white-box analysis, leveraging feature representations at each layer to assess the extent to which the subgraphs formed around each patch remain localized and whether they stay within the object defining the image’s class. To quantify this, we measure visual and embedding-based similarity between interconnected patches, their spatial proximity, and the graph's node separation quality,
correlating these factors with the classification label at different layers. Additionally, we introduce a heatmap-style visualization to track how key patches - particularly those belonging to the main object - relate to their neighbors across layers, offering a more granular view of information flow within the model.
To examine robustness and consistency with human understanding, we not only perform experiments on a subset of the standard ImageNet \cite{imagenet} dataset but also  ImageNet-a \cite{imagenet_o_a}, which contains commonly misclassified adversarial examples. This enables us to explore how the explainability approach holds up when applied to images that may be more challenging to classify, providing deeper insights
in diverse scenarios.

An illustrative example of our approach can be seen in Fig. \ref{fig:teaser}. To elaborate, given an image of dogs in a park, the ViG model correctly classifies it under the "dog" category in ImageNet by progressively representing the image as a dispersed graph of patches, rather than strictly localized pixel groups as CNNs do. Our explainability method provides both visual and numerical insights into this decision. We visualize patch interconnections to examine whether the GNN-based vision model learns to associate \textit{visually} and \textit{semantically} similar patches, regardless of their spatial distance. In Fig. \ref{fig:teaser}, for instance, a green patch corresponding to a dog's face initially connects only to adjacent body parts of the same dog (2nd layer output). However, by the 3rd layer, it becomes linked to more distant red patches corresponding to other dogs' faces. This finding is further reflected in metric diagrams: spatial distance ($D$) increases from layer 2 to layer 3, while visual similarity ($S_{vis}$) and node separation quality ($Q$) decrease. These and additional metrics, detailed in later sections, serve as indicators of the model's decision-making process, ultimately helping to assess how reliably its classifications align with human perception.

In summary, this paper makes the following contributions: 
(1) We introduce the first explainability analysis of GNN-based image classification approaches, 
(2) We propose an evaluation of graph node separation along with semantic, visual, and spatial coherence analysis across layers, combined with heatmap visualizations to track patch interactions, and 
(3) We conduct a comparison of interpretability between standard and adversarial examples to assess robustness and generalizability.

\section{Related Work}
\paragraph{\textbf{Graph Neural Networks in Vision}}

GNNs have gained traction as an alternative to CNNs/ViTs for vision tasks, offering greater flexibility in modeling non-Euclidean relationships between image regions and enhancing efficiency. 
In this direction, ViG \cite{Han2022VisionGA}, introduced a graph-based representation where image patches serve as nodes, with edges encoding semantic and spatial relationships, demonstrating competitive performance against ViTs and CNNs. Feature embeddings are propagated through multiple GNN layers via message passing, and a classification head processes the final node embeddings to predict the final label. 
ViG's innovation includes a \textbf{dynamic graph} update mechanism, contrasting with previous approaches that rely on static superpixel graphs \cite{fey2018splinecnn,avelar2020superpixel}.
Expanding on this, Vision HGNN \cite{Han2023VisionHA} proposed a hypergraph-based approach to capture higher-order feature interactions, further improving recognition accuracy. 
Recent efforts have also focused on enhancing the efficiency of GNNs for mobile applications, as seen in MobileViG \cite{Munir2023MobileViGGS}, which employs sparse attention on a static graph to reduce computational overhead while maintaining high accuracy. 
Additionally, GreedyViG \cite{Munir2024GreedyViGDA} introduces dynamic axial graph construction to optimize graph structure adaptively for further efficiency. 
Such performance-driven design choices lead to more spatially constrained models that are less capable of capturing semantic relationships. As a result, their interpretability is likely to be the same as - or even lower than that of ViG.
Despite their merits, none of these works explore the explainability prospects of their efforts. To address this gap, we focus on ViG 
to investigate how its graph structure can reveal which interconnected image regions contribute to classification, providing visual and quantitative insights into the model’s decision-making process.

\paragraph{\textbf{Image Classification Visualization Techniques}}
A relevant foundational approach follows the use of Class Activation Maps (CAMs) \cite{zhou2016learning}, which highlight regions of an image that significantly influence the model's predictions. 
Building on CAMs, Grad-CAM \cite{Selvaraju2017GradCAM} improved them by enhancing localization, followed by Grad-CAM++ \cite{Chattopadhyay2018GradCAMPlusPlus}, which further refined interpretability. Eigen-CAM \cite{muhammad2020eigen} improved robustness by reducing reliance on gradients. Additionally, \cite{simonyan2013deep} introduced class saliency maps, highlighting influential pixels via gradient-based analysis.
While our approach shares similarities with these well-known techniques, such as highlighting key image regions, we do not aim to directly compare our method to theirs. 
Unlike traditional classifiers, which rely on pixel-level feature extraction, our approach leverages a higher-level GNN backbone, and the nature of the explanation derived from the constructed graph, rather than individual pixels, offers a fundamentally distinct form of interpretability.

\paragraph{\textbf{Explainability in Graph Neural Networks}} Recent research has explored multiple approaches to enhance GNN explainability. 
GNNExplainer \cite{ying2019gnnexplainergeneratingexplanationsgraph} is a post-hoc method which identifies key subgraphs and features through mutual information, while PGExplainer \cite{luo2020parameterizedexplainergraphneural} uses probabilistic masking for instance-level and global explanations.
GraphLIME \cite{huang2020graphlimelocalinterpretablemodel} adapts LIME for localized insights, and PGM-Explainer \cite{vu2020pgmexplainerprobabilisticgraphicalmodel} leverages probabilistic models through perturbations to obtain instance level explanations.
Collectively, these methods advance GNN explainability,
but still leave a significant gap in the explainability of dynamic graph-based models, and specifically to Vision GNNs, which we aim to address.

\section{Background}

\subsection{Notation}

Let $G = (V, E)$ denote a directed graph, where $V = \{v_1, ..., v_N\}$ is the set of $N$ nodes and $E \subseteq V \times V$ is the set of edges. Each node $v_i$ is associated with a feature vector $\mathbf{x}_i \in \mathbb{R}^D$, where $D$ is the feature dimension. The complete set of node features is denoted as $\mathbf{X} = [\mathbf{x}_1,\mathbf{x}_2, ...,\mathbf{x}_N]^\top \in \mathbb{R}^{N \times D}$. The graph structure can be represented by an adjacency matrix $\mathbf{A} \in \mathbb{R}^{N \times N}$, where $A_{ij} = 1$ if there exists an edge from node $i$ to node $j$, and $0$ otherwise. For a node $v_i$, we denote its set of incoming neighbors as $\mathcal{N}(i) = \{j|(j, i) \in E\}$.

We use the superscript 
$l$ to denote layer-specific quantities throughout the network,
where $G^l = (V, E^l)$ represents the graph, $\mathbf{A}^l \in \mathbb{R}^{N \times N}$ the adjacency matrix, and $\mathbf{x}_i^l \in \mathbb{R}^D$ the feature vector of node $i$ at layer $l$. For the input image $\mathbf{I} \in \mathbb{R}^{H \times W \times 3}$, we later partition it into $N = 196$ patches, where each patch $\mathbf{P}_i \in \mathbb{R}^{16 \times 16 \times 3}$ is positioned at coordinates $(r_i, c_i)$ in a $14 \times 14$ grid. For classification outputs, we denote the ground truth class label as $y \in \{1,...,C\}$, the model's output logits as $\hat{\mathbf{y}} \in \mathbb{R}^C$, and the predicted class probabilities after softmax as $\mathbf{p} \in [0,1]^C$.

\subsection{Vision GNN Architecture}

ViG \cite{Han2022VisionGA} processes an input image $\mathcal{I} \in \mathbb{R}^{H\times W\times 3}$ as follows:

\paragraph{\textbf{Patch Embedding}} The image is first resized to $224\times224$ and then partitioned into $N = 14 \times 14 = 196$ non-overlapping patches, where each patch $P_i \in \mathbb{R}^{16\times16\times 3}$ is passed through 2D convolutions to obtain initial node features $\mathbf{x}_i \in \mathbb{R}^D$ (where $D$ is the hidden dimension hyperparameter). Learnable positional encodings $\mathbf{e}_i \in \mathbb{R}^D$ are added to preserve spatial information.

\paragraph{\textbf{Dynamic Graph Construction}} At each layer $l$, a graph $\mathcal{G}^l = (\mathcal{V}, \mathcal{E}^l)$ is constructed by connecting each node to its $K$ nearest neighbors in the embedding space, based on cosine similarity between node features. This yields a layer-specific adjacency matrix $\mathbf{A}^l$ (or adjacency list $\mathcal{E}^l$), which we focus on later.

\paragraph{\textbf{Message Passing}} The proposed Grapher module performs information exchange through message passing. Specifically, the max-relative graph convolution module \cite{Li2019DeepGCNsCG} is chosen for its efficiency, which is formulated as:
\begin{equation}
    \mathbf{x}_i' = \left[ \mathbf{x}_i, max\left( 
\left\{ \mathbf{x}_j-\mathbf{x}_i|j\in \mathcal{N}(\mathbf{x}_i)
 \right\} \right) \right]W_{update}
\end{equation}
where the bias term is omitted. The final update is implemented across multiple heads and the authors apply linear transformations before and after the graph convolution to avoid over-smoothing.

\paragraph{\textbf{Classification}} The final prediction is obtained by applying global average pooling over the node features from the final layer, followed by a linear classifier. For analysis purposes, we can apply this classification head to intermediate layer representations to study the evolution of the model's decision-making process.

\section{Analysis Framework}

\subsection{Datasets}
Our analysis is conducted primarily on ImageNet ILSVRC 2012 \cite{imagenet}, using a subset of 10,000 validation images to establish baseline metrics and patterns. For a more comprehensive understanding of the model's behavior, we extend our analysis to ImageNet-a \cite{imagenet-a-o}, which collects 7,500 natural adversarial images that are commonly misclassified by models that perform competitively on ImageNet.

\subsection{Designed Metrics}
We introduce five quantitative metrics to analyze the evolution of graph structure and decision-making process across layers:

\paragraph{\textbf{1. Embedding Similarity ($S^l_{emb}$)}}: For layer $l$, we compute the average cosine similarity between connected patches:
\begin{equation}
    S^l_{emb} = \frac{1}{|\mathcal{E}^l|} \sum_{(i,j) \in \mathcal{E}^l} \frac{\mathbf{x}^l_i \cdot \mathbf{x}^l_j}{\|\mathbf{x}^l_i\| \|\mathbf{x}^l_j\|}
\end{equation}
where $\mathcal{E}^l$ is the edge set at layer $l$ and $\mathbf{x}^l_i$ represents the embedding of node $i$ at layer $l$. By quantifying the similarity of learned representations between connected patches, we assess if  the model links semantically related regions. Higher values indicate stronger semantic coherence in the graph structure.

\paragraph{\textbf{2. Spatial Distance ($D^l$)}}: We measure the average Manhattan distance between connected patches:
\begin{equation}
    D^l = \frac{1}{|\mathcal{E}^l|} \sum_{(i,j) \in \mathcal{E}^l} (|r_i - r_j| + |c_i - c_j|)
\end{equation}
where $(r_i, c_i)$ represents the grid position of patch $i$. This metric helps us understand if the model maintains local connectivity or gradually forms long-range connections, revealing how the receptive field evolves across layers.

\paragraph{\textbf{3. Visual Similarity ($S^l_{vis}$)}}: For connected patches, we compute the average pixel-space similarity:
\begin{equation}
    S^l_{vis} = \frac{1}{|\mathcal{E}^l|} \sum_{(i,j) \in \mathcal{E}^l} \text{cos}(P_i, P_j)
\end{equation}
where $P_i$ and $P_j$ are the flattened RGB values of patches $i$ and $j$. By comparing the raw RGB values between connected patches, this metric quantifies the level at which the model maintains connections between visually similar regions, therefore tracking how the model transitions from low-level visual features to higher-level semantic representations.

\paragraph{\textbf{4. Layer-wise Classification ($p^l$)}}: For each layer $l$, we compute:
\begin{equation}
    p^l = Pred(\mathbf{X}^l)
\end{equation}
where $Pred$ is the final classification head of the model, and $\mathbf{X}^l$ is the node feature matrix at layer $l$. Only the probability of the ground-truth class $y$ is tracked. This metric reflects how confidence in the ground-truth class evolves across layers.

\paragraph{\textbf{5. Object-based Graph Modularity ($Q^l$)}}: Given a binary mask $M$ indicating the image patches that include the ground-truth object, we compute the graph modularity score:
\begin{equation}
     Q^l = \sum_{c=1}^{2} \left[ \frac{L_c}{|\mathcal{E}^l|} - \left( \frac{k^{in}_c k^{out}_c}{2|\mathcal{E}^l|} \right)^2 \right]
\end{equation}
where $L_c$ is the number of intra-community links for each community $c$ (object or background), $k^{in}_c$ and $k^{out}_c$ are the sums of incoming and outgoing degrees of the nodes in a community, respectively. For this metric, we start by partitioning the graph into two sets of patches, those that belong to the main object, and those that do not. To automate this pipeline, we utilize GroundingDino \cite{liu2024grounding} along with SAM \cite{kirillov2023segment}, to compute the object binary masks. Then, we use graph modularity, to measure how much the edges of the graph 
stay localized within each group, or cross between them. This is a particularly insightful metric in our use case, since it allows us to monitor how the edges' separation quality of the generated graph evolves, across different layers.

\paragraph{\textbf{Visual Explanation through Connection Heatmaps}}: 
To qualitatively analyze the model's attention patterns, we visualize the connection structure for specific patches across different layers. For a selected patch (green), we create a heatmap showcasing the generated incoming edges of this patch (red), where the intensity represents the embedding-based similarity score between connected patches. 
As demonstrated in Fig. \ref{fig:extra-qual}, this approach reveals how a selected patch on the ship's hull (green) forms connections with varying intensities of red, where brighter red indicates stronger embedding similarity. This visualization aids in understanding how the model gradually builds connections between different regions of the image.

\section{Results}

\subsection{Quantitative Results}

Results correspond to experiments conducted on the ViG-Small variant of the Vision GNN \cite{Han2022VisionGA}. Across all three datasets consistent patterns emerge in how the model's graph structure evolves through the intermediate layers. In the standard ImageNet validation set (Table \ref{tab:imagenet-stacked-vertical}), we observe a clear trade-off between local and global feature integration. Early layers (1-4) maintain high visual ($S^l_{vis} > 0.6$) and embedding similarities ($S^l_{emb} > 0.9$) with relatively short spatial distances ($D^l < 4$), indicating a focus on \textbf{local feature extraction}. As the network deepens, these metrics gradually shift, with spatial distances increasing significantly ($D^l \approx 8.8$ in final layers) while visual similarity decreases ($S^l_{vis} \approx 0.3$), 
indicating a more explorative behavior, that also increases the effective receptive field of each patch.
The emergence of class-specific representations in deeper layers is supported by the sudden increase in embedding similarity ($S^l_{emb}$) during the final two layers, suggesting a convergence towards more \textbf{semantically meaningful features}. This evolution correlates with improving classification accuracy, reaching $68.6\%$ in the final layers.
\begin{table}[t]
\centering
\caption{Analysis metrics across model layers on \textbf{ImageNet} (top) and \textbf{ImageNet-a} (bottom). Top three values for each dataset per metric are indicated by \textbf{bold} (highest), $^\star$ (second), and $^\dagger$ (third).}
\vspace{6pt}
\label{tab:imagenet-stacked-vertical}
% \begin{tabular}{c|c|cccccc}
\begin{tabular}{c|>{\centering\arraybackslash}p{1cm}|>{\centering\arraybackslash}p{1.1cm}>{\centering\arraybackslash}p{1.1cm}>{\centering\arraybackslash}p{1.1cm}>{\centering\arraybackslash}p{1.1cm}>{\centering\arraybackslash}p{1.1cm}>{\centering\arraybackslash}p{1.8cm}}
\toprule
 & Layers & $S^l_{vis}$ & $D^l$ & $S^l_{emb}$ & $Q^l$ & $p^l$ & Top-1 Acc.\\
\midrule
\multirow{8}{*}{\rotatebox{90}{\textbf{ImageNet}}}
 & 1-2   & \textbf{0.700} & 3.525 & \textbf{0.949} & \textbf{0.236} & 0.005 & 0.008 \\
 & 3-4   & 0.640$^\star$ & 3.672 & 0.936$^\star$ & 0.234$^\star$ & 0.009 & 0.018 \\
 & 5-6   & 0.556$^\dagger$ & 5.394 & 0.908$^\dagger$ & 0.204$^\dagger$ & 0.019 & 0.045 \\
 & 7-8   & 0.513 & 5.848 & 0.889 & 0.186 & 0.049 & 0.133 \\
 & 9-10  & 0.451 & 6.593 & 0.854 & 0.160 & 0.118 & 0.281 \\
 & 11-12 & 0.421 & 7.142$^\dagger$ & 0.842 & 0.146 & 0.244$^\dagger$ & 0.449$^\dagger$ \\
 & 13-14 & 0.357 & 8.160$^\star$ & 0.845 & 0.119 & 0.379$^\star$ & 0.600$^\star$ \\
 & 15-16 & 0.306 & \textbf{8.858} & 0.900 & 0.088 & \textbf{0.486} & \textbf{0.686} \\
\midrule
\multirow{8}{*}{\rotatebox{90}{\textbf{ImageNet-a}}}
 & 1-2   & \textbf{0.700} & 3.537 & \textbf{0.953} & 0.095$^\star$ & 0.001 & 0.000 \\
 & 3-4   & 0.643$^\star$ & 3.651 & 0.940$^\star$ & \textbf{0.096} & 0.001 & 0.001 \\
 & 5-6   & 0.563$^\dagger$ & 5.398 & 0.912$^\dagger$ & 0.079$^\dagger$ & 0.001 & 0.002 \\
 & 7-8   & 0.521 & 5.890 & 0.893 & 0.071 & 0.002 & 0.003 \\
 & 9-10  & 0.463 & 6.638 & 0.859 & 0.058 & 0.003 & 0.004 \\
 & 11-12 & 0.434 & 7.165$^\dagger$ & 0.846 & 0.052 & 0.004$^\dagger$ & 0.007$^\dagger$ \\
 & 13-14 & 0.371 & 8.128$^\star$ & 0.845 & 0.040 & 0.007$^\star$ & 0.012$^\star$ \\
 & 15-16 & 0.320 & \textbf{8.845} & 0.877 & 0.030 & \textbf{0.018} & \textbf{0.027} \\
\bottomrule
\end{tabular}
\end{table}

This progressive transformation of the graph structure closely mirrors the hierarchical feature learning observed in traditional CNNs. Similar to how CNN kernels gradually expand their receptive field to capture increasingly complex patterns, ViG systematically builds connections across larger spatial distances while transitioning from low-level visual features to more abstract class-specific representations.
Compared to the fixed kernel sizes of a CNN, however, this dynamic graph construction offers an inherently interpretable view of how the model processes information - each edge in the graph explicitly reveals which image regions the model considers relevant for feature extraction, effectively providing a self-documenting receptive field that adapts to the input content. 

When comparing the metrics between ImageNet validation and ImageNet-a datasets (Table \ref{tab:imagenet-stacked-vertical}), the most striking difference appears in the graph modularity scores ($Q^l$), which are significantly lower for adversarial images (starting at 0.095 compared to 0.236 for standard images). This suggests that the model struggles to maintain coherent graph structures that separate object regions from the background in challenging cases. This degradation in graph structure is reflected in the dramatically reduced classification performance, with top-1 accuracy dropping from 68.6\% to just 2.7\% and correct class probability ($p^l$) falling from 0.486 to 0.018 in the final layers. While the model's performance shows that it still encounters the same pitfalls as models trained on large-scale datasets (strong intra-domain performance but limited generalization ability) we showcase that our framework still reliably provides insights even in failure cases.

\begin{figure}[t]
    \centering
    \includegraphics[width=0.95\linewidth]{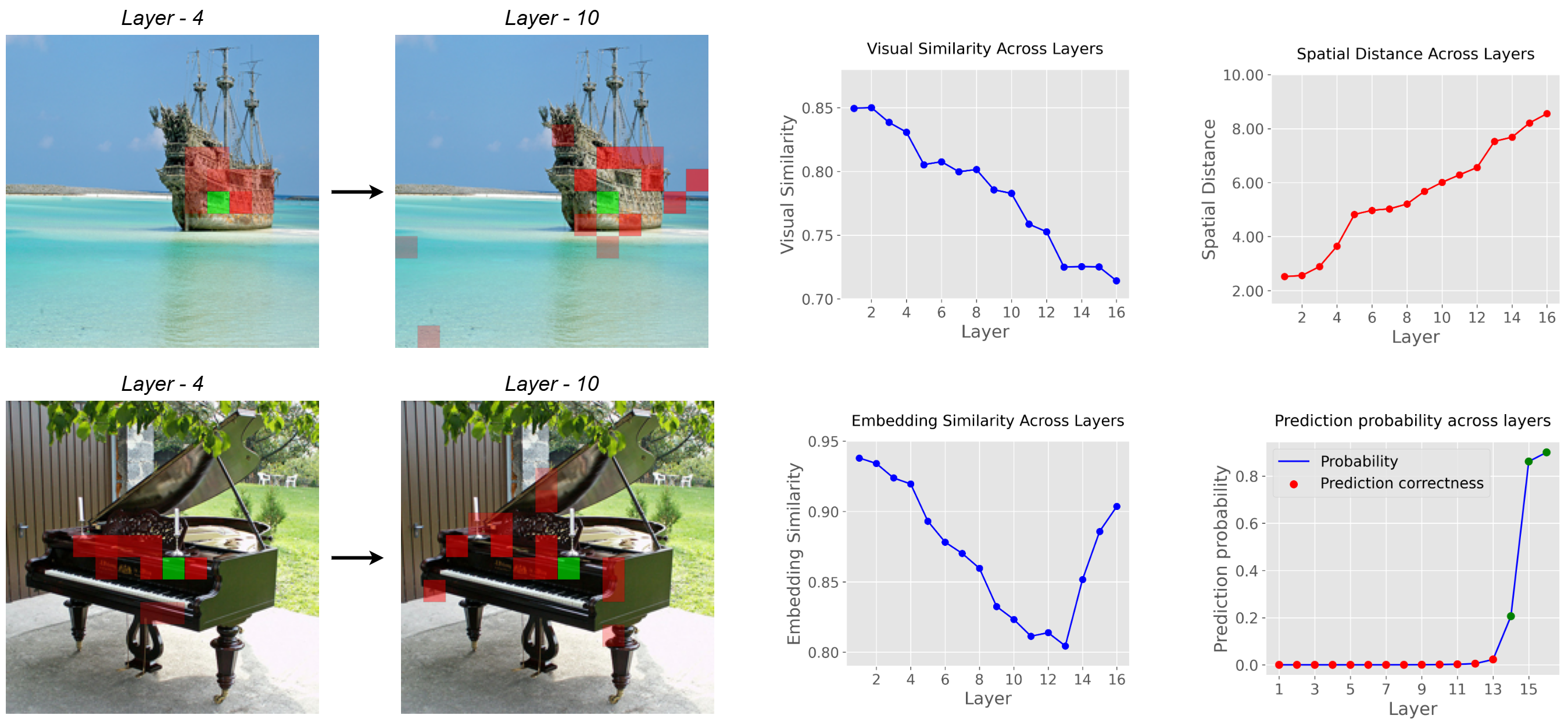}
    \caption{Heatmap Visualization of intermediate graphs (layers 4 and 10), for two ImageNet images, and metric evolution across all layers Visual Similarity, Spatial Distance (top image) and Embedding Similarity and Prediction probability (bottom image).}
    \label{fig:extra-qual}
\end{figure}

\subsection{Qualitative Results}

To complement our quantitative analysis, we visualize the evolution of graph connections across selected cases and interpret the results using our proposed metrics. As shown in Fig. \ref{fig:extra-qual}, examining the top image of a ship reveals how the model's connection patterns differ from layer 4 to layer 10. In the earlier layer, connections (shown in red) from the selected patch (in green) remain largely concentrated within the main object's structure, maintaining high visual similarity with neighboring patches (indicated by the intensity of the red patches). By layer 10, these connections extend further spatially, reaching into the background regions, while maintaining stronger similarities with patches that belong to the ship. This aligns with our quantitative observations in the adjacent diagrams that showcase increasing spatial distance and decreasing visual similarity as we observe deeper layers in the model. The bottom example of a piano further illustrates this progression while also demonstrating a particularly interesting pattern in the final layers - the sharp increase in prediction probability coincides with a notable spike in embedding similarity during the last three layers. This correlation supports our earlier quantitative findings about the model's convergence toward class-specific representations in deeper layers, suggesting a crucial phase where the model consolidates its classification decision. Notably, this critical decision-making phase occurs precisely when visual similarity between connected patches is at its lowest, revealing that the model's most confident predictions emerge from representations that diverge significantly from human visual intuition.

\begin{figure}[t]
    \centering
    \begin{minipage}{0.65\textwidth}
        \centering
        \subfloat[]{\includegraphics[width=\textwidth]{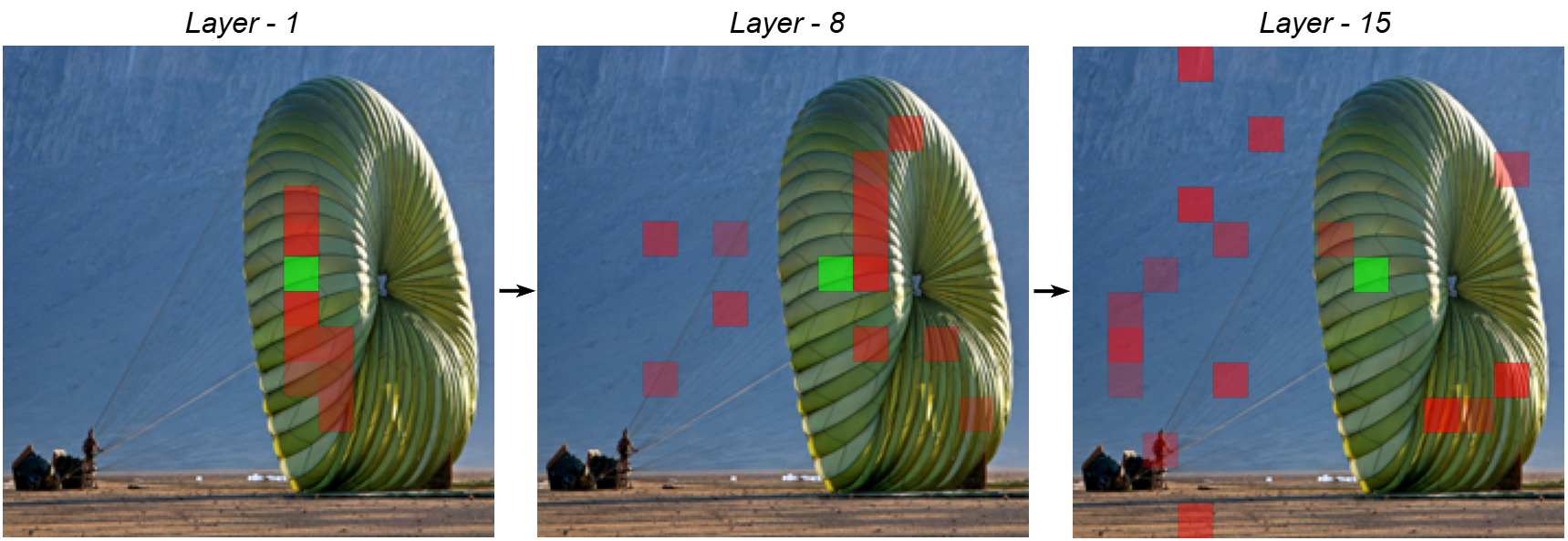}}\\
        \subfloat[]{\includegraphics[width=\textwidth]{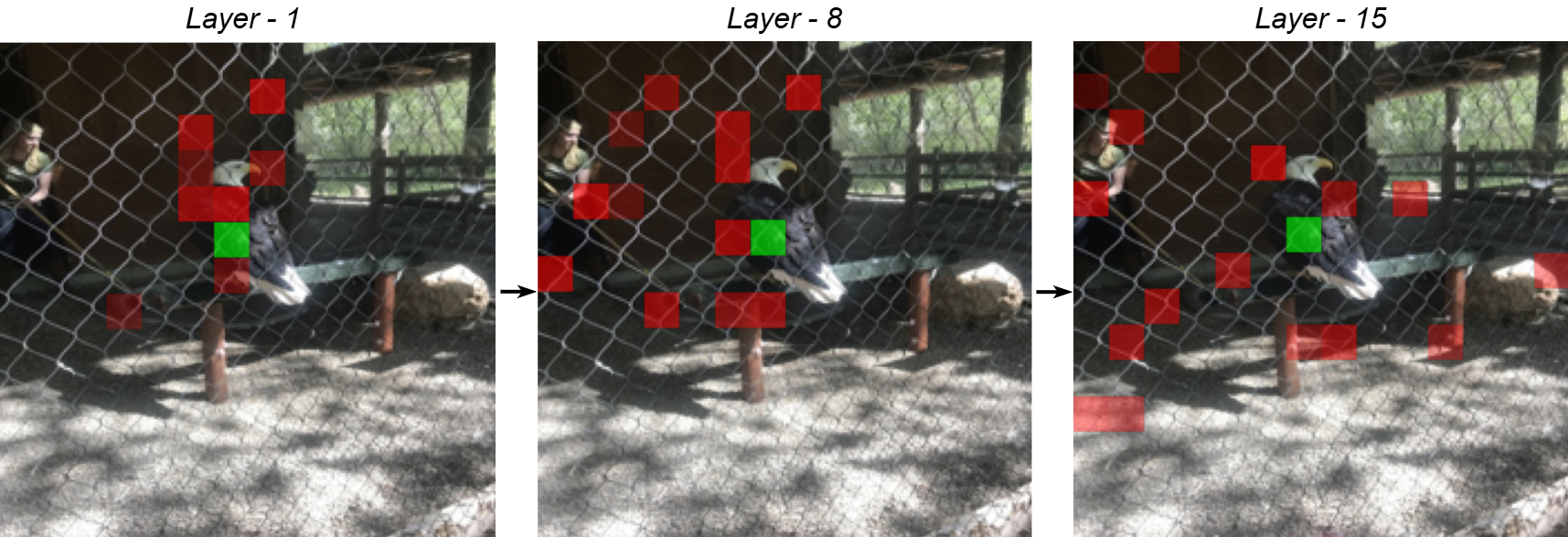}}
    \end{minipage}\hfill
    \begin{minipage}{0.35\textwidth}
        \centering
        \subfloat[]{        \includegraphics[width=0.85\textwidth]{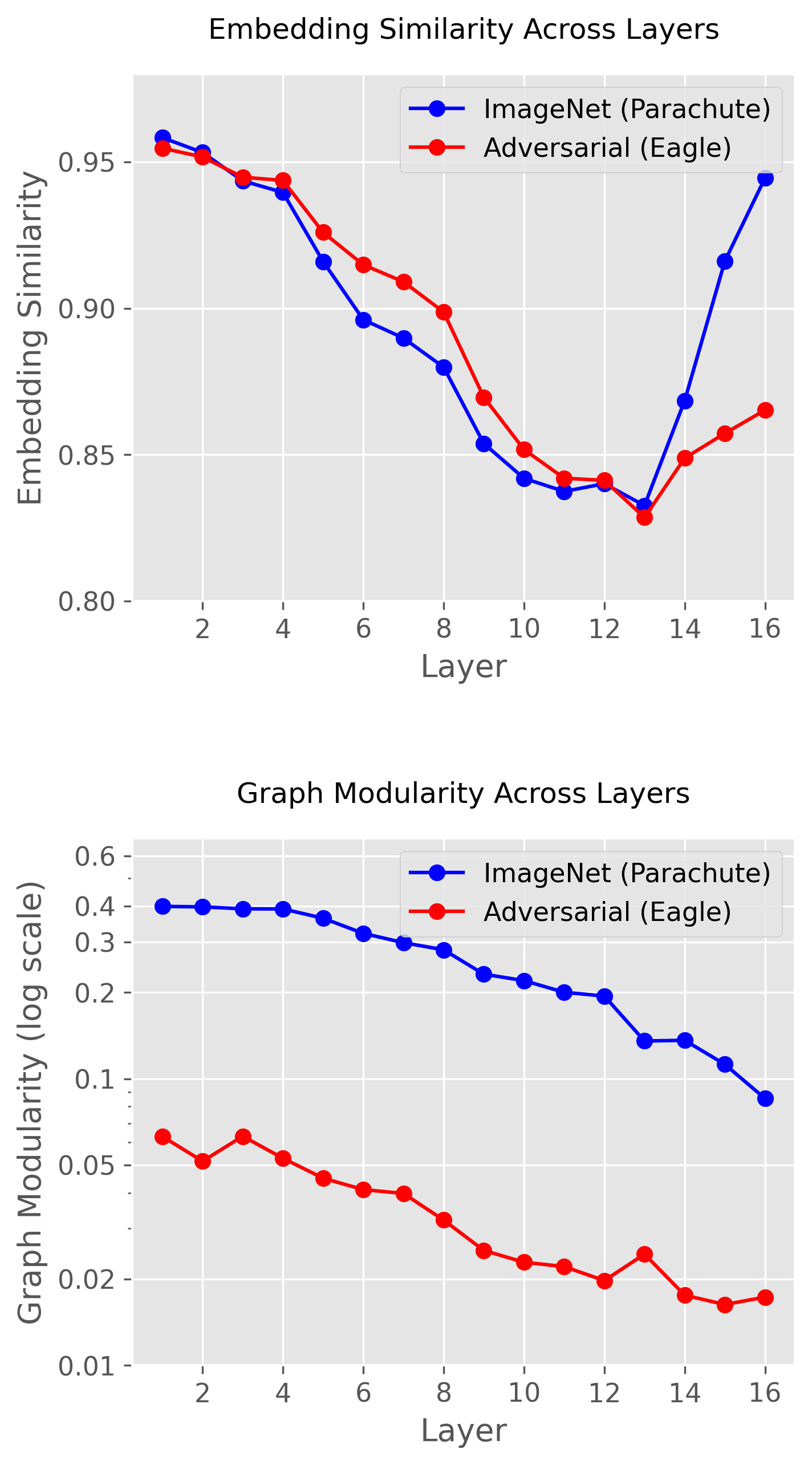}}

    \end{minipage}
    
    \caption{Heatmap Visualization of intermediate graphs (layers 1, 8 and 15), for intra-domain ImageNet (a) and adversarial ImageNet-a (b) images, and metric evolution across all layers for Embedding Similarity and Graph Modularity (c).}
    \label{fig:main-qual}
    \vskip -0.2in
\end{figure}

Comparing the in-domain ImageNet image (Fig. \ref{fig:main-qual}(a)) with an adversarial image from the ImageNet-a dataset (Fig. \ref{fig:main-qual}(b)) reveals important differences in how the model builds and maintains connections. The adversarial example shows significantly more dispersed connectivity patterns even in early and middle layers, with connections constantly extending beyond the main object. This behavioral difference is quantitatively captured in the Graph Modularity diagram (Fig. \ref{fig:main-qual}(c)), where the adversarial example shows consistently lower modularity scores across all layers, indicating weaker separation between object and background regions. Furthermore, the adversarial image is misclassified as a dog breed, which is also reflected in the Embedding Similarity diagram. In this diagram, the characteristic spike in similarity during the final layers
(associated with convergence to class-specific features, as shown in the Quantitative section)
is notably diminished in the adversarial case. While the in-domain example shows a sharp increase in embedding similarity in layers 15-16 (reaching ~0.95), the adversarial case exhibits a much more modest increase (only reaching ~0.86), indicating potential difficulties in forming consistent class-specific representations.

\section{Conclusion}
In this work, we presented the first comprehensive explainability analysis of graph-based image classification with Vision GNNs, proposing novel quantitative metrics and visualization techniques to understand how these models process and encode visual information. Our analysis revealed that the model exhibits a clear progression from local to global feature processing across layers, demonstrated by increasing spatial distances and evolving similarity patterns between connected patches. Through experiments on both the standard ImageNet dataset and adversarial variations, we showed that our proposed metrics and visualization techniques can provide valuable insights into the model's behavior and decision-making, even in cases where classification fails. 
However, our approach has certain limitations, such as its white-box access of specific model architectures and potential computational overhead. Future research should explore how this method can be further leveraged to enhance the interpretability and performance of Vision GNNs, potentially guiding architectural improvements and integrating these metrics into a re-training stage to refine model performance.

\begin{credits}
\subsubsection{\ackname} This work was supported by the Hellenic Foundation for Research and Innovation (HFRI)  under the 5th Call for HFRI PhD Fellowships
(Fellowship Number 19268). We acknowledge the use of the Amazon Web Services
(AWS) platform for providing the infrastructure to build and test our experimental setup.

\subsubsection{\discintname}
The authors have no competing interests to declare that are
relevant to the content of this article. 
\end{credits}


\bibliographystyle{splncs04}
\bibliography{main}
\end{document}